\documentclass{article}
\usepackage{ijcai23}

\usepackage{times}
\usepackage{soul}
\usepackage{url}
\usepackage[hidelinks]{hyperref}
\usepackage[utf8]{inputenc}
\usepackage[small]{caption}
\usepackage{graphicx}
\usepackage{amsmath}
\usepackage{amsthm}
\usepackage{booktabs}
\usepackage{algorithm}
\usepackage{algorithmic}
\usepackage[switch]{lineno}

\usepackage{tabularx}
\usepackage{array}
\usepackage{makecell}
\usepackage{caption}
\usepackage{subcaption}
\usepackage{xcolor}
\usepackage{latexsym}
\usepackage{amssymb}
\usepackage{bm}
\usepackage{multirow}
\newcommand\red[1]{\textcolor{red}{#1}}

\newcommand\green[1]{\textcolor[RGB]{0,100,0}{#1}}

\usepackage[T1]{fontenc}

\title{Decoding the Underlying Meaning of Multimodal Hateful Memes}

\author{
Ming Shan Hee$^1$
\and
Wen-Haw Chong$^2$
\And
Roy Ka-Wei Lee$^1$
\affiliations
$^1$Singapore University of Technology and Design\\
$^2$Singapore Management University
\emails
mingshan\_hee@mymail.sutd.edu.sg,
whchong.2013@phdis.smu.edu.sg,
roy\_lee@sutd.edu.sg
}

\begin{document}

\maketitle

\begin{abstract}
Recent studies have proposed models that yielded promising performance for the hateful meme classification task. Nevertheless, these proposed models do not generate interpretable explanations that uncover the underlying meaning and support the classification output. A major reason for the lack of explainable hateful meme methods is the absence of a hateful meme dataset that contains ground truth explanations for benchmarking or training. Intuitively, having such explanations can educate and assist content moderators in interpreting and removing flagged hateful memes. This paper address this research gap by introducing \textbf{Hat}eful meme with \textbf{Re}asons \textbf{D}ataset (\textsf{HatReD}), which is a new multimodal hateful meme dataset annotated with the underlying hateful contextual reasons. We also define a new conditional generation task that aims to automatically generate underlying reasons to explain hateful memes and establish the baseline performance of state-of-the-art pre-trained language models on this task.  We further demonstrate the usefulness of \textsf{HatReD} by analyzing the challenges of the new conditional generation task in explaining memes in seen and unseen domains. The dataset and benchmark models are made available here: \url{https://github.com/Social-AI-Studio/HatRed}
\end{abstract}

\textcolor{red}{Disclaimer: This paper contains discriminatory content that may be disturbing to some readers.}

\section{Introduction}
\label{sec:intro}
\textbf{Motivation.} Internet memes are viral content spread among online communities. While most memes are often humorous and benign, hateful memes, which attack a target group or individual based on characteristics such as race, gender, and religion, have become a growing concern. As part of its effort to moderate the spread of hateful memes, Facebook recently launched the ``\textit{hateful meme challenge}''~\cite{kiela2020hateful}. The challenge released a dataset with 10K+ hateful memes to encourage submissions of automated solutions to detect hateful memes. This led to the development of various multimodal deep learning approaches for hateful meme classifications ~\cite{yang2023crossdomain,models-lee-2021-disentangling}. Other studies have also contributed to the growing effort to curb hateful memes by collecting and releasing large hateful meme datasets to support the training and evaluation of hateful meme classification models ~\cite{dataset-suryawanshi-2020-multioff,dataset-gasparini-2021-misogyny,dataset-pramanick-2021-harmful,sharma2023characterizing,sharma2022detecting}.

\begin{figure}[t]
    \centering
    \includegraphics[scale=0.525]{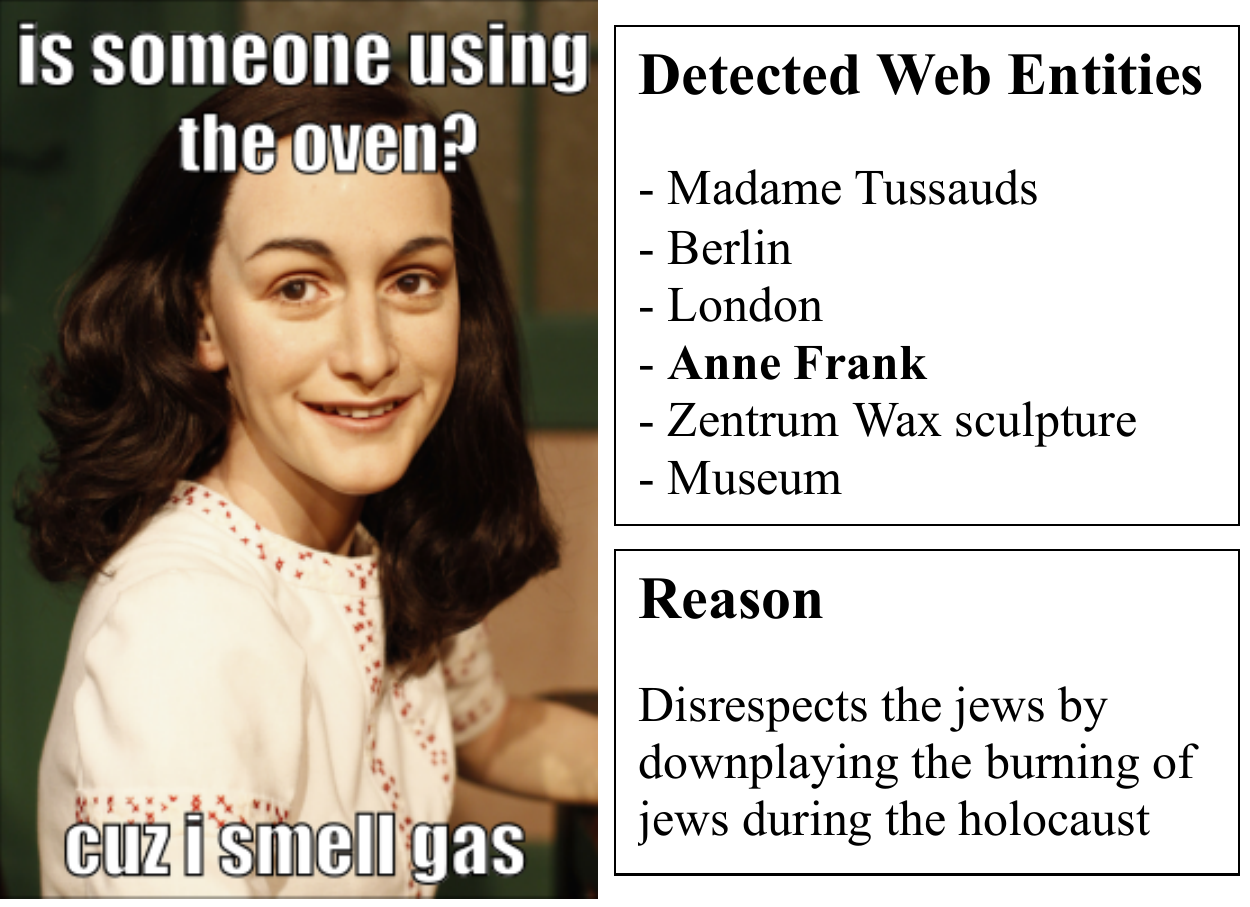}
    \caption{Example of a hateful meme in \textsf{HatReD}.}
    \label{fig:explanation-sample-meme}
\end{figure}

However, existing studies have primarily focused on performing hateful meme classification (i.e., predicting if a given meme is hateful) with limited explanations for its prediction. Providing explanations for the detected hate meme is integral to the content moderation process. The content moderators and users may want to understand why a particular meme is flagged as hateful. Nevertheless, explaining hateful memes is challenging as it requires a combination of information from various modalities and specific socio-cultural knowledge ~\cite{kiela21a-hatefulmemes}. Consider the hateful meme in Figure~\ref{fig:explanation-sample-meme}. To explain the hateful meme, one will need the socio-cultural knowledge that the girl in the image is Anne Frank, and realize the textual reference refers to the gas poisoning of Jews during the Holocaust. 

Recognizing the importance of providing contextual reasons for the predicted hateful memes, recent studies have performed fine-grained analysis to classify the type of attacks \cite{dataset-mathias-2021-findings} and infer the targets being attacked \cite{dataset-pramanick-2021-harmful,sharma2022disarm}. However, such fine-grained analysis may still be inadequate for content moderators to understand hateful memes. These analyses often only predict general protected characteristics (e.g., race) but not the specific social target attacked (e.g., Jews). Furthermore, having informative reasons in natural sentences, such as the example provided in Figure~\ref{fig:explanation-sample-meme}, would make it easier for content moderators to comprehend the hateful memes.

In a recent study, Elsherief \shortcite{elsherief-2021-latent} collected a large textual dataset to support implicit hate speech classification. The dataset contains hateful textual posts annotated with their corresponding \textit{implied statements}, which could be seen as a form of explanation to aid content moderators in understanding the implicit hate speech. The availability of ground truth explanations also allows researchers to apply and explore training Pre-trained Language Models (PLMs) such as GPT-2 for explanation generation. Ideally, we would also like to generate the underlying reasons for why a flagged hateful meme is considered hateful. To the best of our knowledge, there is no current dataset to facilitate this exploration.

\paragraph{Research Objectives.} To address the research gaps, we propose a new conditional generation task that aims to generate the underlying reasons to explain hateful memes automatically. We constructed the \textbf{Hat}eful meme \textbf{Re}asoning \textbf{D}ataset (\textsf{HatReD}), which is a new multimodal hateful memes dataset annotated with the underlying hateful contextual reasons, to support the proposed task. Specifically, we carefully design a framework to annotate Facebook's \textit{Fine-Grained Hateful Memes} dataset~\cite{dataset-mathias-2021-findings} with the underlying hateful reasons. We fine-tune PLMs on \textsf{HatReD} and conduct extensive experiments to evaluate the PLM's performance and limitations on the new generation task. Finally, we also demonstrate the usefulness of \textsf{HatReD} by evaluating the fine-tuned PLMs' ability to generate the explanations for hateful memes in an unseen misogynous meme dataset.

\paragraph{Contributions.} We summarize our contributions as follows: (1) We construct \textsf{HatReD}, which is a multimodal hateful meme dataset annotated with underlying hateful contextual reasons. To the best of our knowledge, this is the first hateful meme dataset with written explanations. (2) We introduce a new conditional generation task that aims to generate underlying reasons to explain hateful memes automatically. We conduct extensive experiments to establish the task baselines using state-of-the-art PLMs. (3) We analyze the challenges of generating the generation new task and demonstrate the usefulness of \textsf{HatReD} in explaining memes in seen and unseen domains.

\section{Related Works}
\label{sec:related}

\begin{table*}[ht]
\centering
\small
\begin{tabular}{llrrccc}
\toprule
Work & Domain & Size & Num. Hateful/Off. & Target/Group & Attack Type & Explanations \\
\midrule
Kiela \shortcite{kiela2020hateful} & Multiple Groups & 10000 & $3,266^{*}$ & & & \\ 
Suryawanshi \shortcite{dataset-suryawanshi-2020-multioff} & Politics & 743 & 305 & & & \\ 
Mathias \shortcite{dataset-mathias-2021-findings} & Multiple Groups & 10,000  & $3,253^{*}$ & \checkmark & \checkmark & \\ 
Pramanick \shortcite{dataset-pramanick-2021-harmful} & COVID-19 & 3,544  & 1,249  & \checkmark &  & \\ 
Gasparini \shortcite{dataset-gasparini-2021-misogyny} & Misogyny & 800  & 400 & &  & \\ 
Fersini \shortcite{fersini2022semeval} & Misogyny & 11,000 & $5,504^{*}$  & &  & \\ 
HatReD (Ours) & Multiple Groups & 3,228 & 3,228  & \checkmark & \checkmark & \checkmark \\ 
\bottomrule
\end{tabular}
\caption{Summary of hateful meme datasets.All the datasets contain class labels that support the hateful meme classification task. However, none of the existing datasets provide explanation for the hateful context. \textsf{HatReD} is the first dataset that include free-text explanations for multi-modal memes in Hearst-style templates. * indicates that the hateful and/or offensives memes in the test set are excluded, as the test set are not made publicly available.
}
\label{tab:dataset-comparisons-table}
\end{table*}

\subsection{Hateful Meme Datasets}
Hateful meme classification is an emerging research topic made popular by the availability of several hateful meme datasets. Table~\ref{tab:dataset-comparisons-table} summarizes the hateful meme datasets released over the last few years. All the datasets contain class labels that support the hateful meme classification task. For instance, the memes in \textit{Facebook Hateful Meme Challenge} dataset are labeled ``\textit{hateful}'' or ``\textit{non-hateful}'' \cite{kiela2020hateful}. Similarly, Suryawanshi \shortcite{dataset-suryawanshi-2020-multioff} collected a small dataset of politics-related memes from Tumblr and annotated the memes as ``\textit{offensive}'' or ``\textit{non-offensive}''. Besides the class labels that facilitate hateful meme classification, some datasets have also provided supplementary information on hateful memes. For example, Pramanick \shortcite{dataset-pramanick-2021-harmful} collected a dataset containing COVID-19 related memes. The researchers annotated the harmfulness of the memes and the types of target (e.g., \textit{individual}, \textit{organization}, and \textit{community}) attacked in the harmful memes. Mathias \shortcite{dataset-mathias-2021-findings} extended the Facebook Hateful Meme Challenge dataset by annotating the types of attack (e.g., \textit{Dehumanizing}) and target type (e.g., \textit{race}) attacked in the hateful memes. While the supplementary information could provide additional contexts to the hateful memes, it is still inadequate in informatively explaining the hateful memes.


Recent studies have attempted to identify and explain the subtle hateful connotations of hate speech. Sap \shortcite{sap-2018-sbf} developed the \textit{Social Bias Frame}, a pragmatic framework that can capture knowledge regarding the biased implications of hate speech, such as its group reference and implied statement. Elsherief \shortcite{elsherief-2021-latent} subsequently extended the \textit{Social Bias Frame} to include implicit hate speech, which has a broader scope than social bias and stereotypes. Nevertheless, these existing studies have mainly focused on explaining text-based hate speeches. In this study, we aim to fill the research gap by proposing a new hateful meme dataset that includes informative reasons to explain the background contexts in hateful memes.

\subsection{Hateful Meme Classification}
Hateful meme classification is an emerging multimodal task that has gain popularity in recent years. Existing studies have explored \textit{classic two-stream models} that combine the text and visual features to classify the hateful memes~\cite{kiela2020hateful,dataset-suryawanshi-2020-multioff}, and fine-tuning large scale pre-trained multimodal models for the multimodal classification task~\cite{lippe2020multimodal,DBLP:journals/corr/abs-2012-07788,DBLP:conf/emnlp/PramanickSDAN021,yang2023crossdomain,models-lee-2021-disentangling,cao2023prompting,sharma2022detecting}. Nevertheless, most of the existing studies have focused on the hateful meme classification task without providing any explanation for the hateful memes. A recent study proposed a post-hoc explanation framework to examine the visual-text slur grounding learned by pre-trained multimodal models trained to perform the hateful meme classification task \cite{2022-hee-explaining}. However, the framework still falls short in providing informative reasons to explain hateful memes. We postulate that the primary reason for underwhelming research studies explaining hateful memes is the lack of a dataset. Therefore, we propose a multimodal hateful meme dataset with informative reasons to encourage researchers to contribute solutions in this space. Specifically, this study will provide the benchmark dataset for the hateful meme explanation task and comprehensively evaluate state-of-the-art PLMs' capabilities to generate natural language reasons for hateful memes.


\section{HatReD Dataset}
\label{sec:annotation}
In this study, we propose \textsf{HatReD}\footnote{Note that researchers will have to agree with Facebook's data access agreement to download the memes}, a new multimodal hateful meme explanation dataset. Specifically, we recruited four native English speakers to annotate the underlying reasons for the hate speeches found in Facebook's \textit{Fine-Grained Hateful Memes} dataset \cite{dataset-mathias-2021-findings}. To the best of our knowledge, this is the first multimodal hateful meme dataset annotated with hateful contextual reasons. In the subsequent sections, we will discuss the dataset construction process and provide a preliminary analysis of \textsf{HatReD}.

\subsection{Dataset Construction}
\label{sec:data_construct}

\textit{Fine-Grained Hateful Memes} is a large-scale multimodal memes benchmark dataset that contains five standard kinds of incitement to hatred, including sexual, racial, religious, nationality and disability hatred. The selection of these curated memes conforms with the community standards on hate speech employed by Facebook \footnote{https://www.facebook.com/communitystandards/hate\_speech}, which presents a pragmatic view of hate speech in memes.

\subsubsection{Dataset and Annotation Preparation} The main challenge of explaining hateful memes is that the explanation often requires knowledge of relevant socio-cultural backgrounds and societal prejudices. For example, recognizing that the presence of rainbow-striped flags may indicate a connection to LGBTQ movement, such as Pride Day. Therefore, to assist annotators in explaining hateful memes, we used the Google Web Detect API to extract web entities from the meme. The extracted web entities could provide the additional socio-cultural context for the images used in the memes. For instance, the API will return ``\textit{Anne Frank}'' for the image used in Figure~\ref{fig:explanation-sample-meme}. The annotators are also encouraged to search about the unfamiliar extracted web entities and slurs in meme text on external knowledge bases such as Wikipedia and Hatebase\footnote{https://hatebase.org/}. For instance, annotators without prior knowledge of \textit{Anne Frank} can search Wikipedia for information about her and events related to her, such as the Holocaust. Through this process, the annotators can deepen their knowledge of relevant cultural backgrounds and societal prejudices over multiple iterations of annotation and improve their annotation.


\begin{table}[t]
\centering
\small
\begin{tabular}{ccc}
    \toprule
    & \textbf{Fluency} & \textbf{Relevance} \\
    \midrule
    \textbf{Average Score} & \textit{4.97} & \textit{4.81} \\
    \bottomrule
    \hline
\end{tabular}
\caption{Human evaluation results on annotated reasons}
\label{tab:human-evaluation-train-dataset}
\end{table}

\begin{table}[t]
    \centering
    \small
    \begin{tabular}{lcc}
    \toprule
    \textbf{Social Category} & \textbf{\# Social Targets} & \textbf{\# Reasons} \\
    \midrule
    Sex & 3 & 673 \\
    Race & 7 & 884 \\
    Religion & 8 & 1,188 \\
    Nationality & 34 & 328 \\
    Disability & 2 & 231 \\
    \midrule
    \textbf{\textit{total}} & \textit{54} & \textit{3,304} \\
    \bottomrule
    \end{tabular}
    \caption{The distribution of social targets and annotated reasons within each social category in \textsf{HatReD}.}
    \label{tab:social-targets-statistic}
\end{table}

\subsubsection{Reason Annotation} We trained the four annotators to produce high-quality reasons for each hateful meme. We present annotators with the meme, the social characteristics of the attacked target (e.g., nationality), the type of attack (e.g., contempt), and the extracted web entities. Annotators have to identify and explain the hate speech with three primary goals:  (i) the annotated reasons should specify and cover all implied hate speeches in the meme, (ii) the annotated reasons should accurately express and reflect the underlying hate implication, and (iii) the annotated reasons should be fluent and grammatically correct. We also ensure the annotated reasons are consistent across the annotators by requesting the reasons to be written in one of the two following Hearst-like patterns: (i) \textit{<verb> <target> <predicate>} or (ii) \textit{use of derogatory terms against <target> <predicate>}, where \textit{<target>} represents the attacked social target and \textit{<predicate>} highlights the hateful implication. 

\subsubsection{Annotation Quality Control} We conducted four trial annotations to ensure that the annotators were competent and proficient for the task. In each trial, 20 unique hateful memes are sampled for each annotator, and the annotators are tasked to craft the hateful reasons. At the end of each trial, the annotators will assess the quality of hateful reasons written by other annotators. The annotated reasons are evaluated based on the following criteria:

\begin{itemize}
    \item \textit{Fluency}: Rate the structural and grammatical correctness of the reasons using a 5-point Likert scale. 1: unreadable reasons with too many grammatical errors, 5: well-written reasons with no grammatical errors.
    \item \textit{Relevance}: Rate the relevance of the reasons using a 5-point Likert scale. 1: reasons misrepresent the implied hate speech, 5: reasons accurately reflect the implied hate speech
\end{itemize}

At the end of each iteration, we present the evaluation ratings of their hateful reasons to the annotators and discuss how to improve the poorly rated hateful reasons. These discussions helped our annotators to improve the quality of their annotation.

\subsection{Corpus Analysis}

In total, \textsf{HatReD} dataset contains 3,304 annotated reasons for 3,228 hateful memes. Some memes may have multiple annotated reasons because they attack multiple social groups. The minimum explanation length is 5, the mean explanation length is 13.62, and the maximum is 31.

To examine the quality of the annotated reasons, we conducted human evaluation similar to our annotation trial on 1,200 hateful memes in our \textsf{HatReD} dataset. The annotators are tasked to evaluate the annotated reasons written by others. This translates into having three human evaluation results for each annotated reason. Table \ref{tab:human-evaluation-train-dataset} shows the average \textit{fluency} and \textit{relevance} of the human evaluation on the hateful reasons for 1,200 memes. We observe a high average score of 4.97 and 4.81 for \textit{fluency} and \textit{relevance}, respectively, suggesting that the annotated reasons fluently capture the hateful speeches in the memes. Furthermore, we observe that the three evaluators have a unanimous agreement for \textit{fluency} ratings in 93.9\% of the evaluated annotated reasons, i.e., the evaluators rated the same score in 93.9\% of the evaluated reasons. Similarly, the evaluators unanimously agree for their \textit{relevance} ratings in 81.2\% of the evaluated annotated reasons. 

Table \ref{tab:social-targets-statistic} illustrates the distribution of the social targets and hateful reasons found in the \textsf{HatReD} dataset. We observe significant variations in the number of social targets per social category. For example, there are 673 memes targeting the \textit{Sex} social group, comprising three social targets (i.e., \textit{LGBT}, \textit{Female}, and \textit{Males}). In contrast, the \textit{Nationality} social group has 328 memes are attacking 34 unique social targets. Therefore, we expect more diverse and sparsely annotated reasons for hateful memes targeting \textit{Nationality} social targets. 

\section{Hateful Memes Explanation}
\label{sec:task}
The availability of \textsf{HatReD} enables the exploration of a new task, \textit{hateful meme explanation}. Specifically, we propose a natural language generation task where trained models generate the underlying reasons to explain hateful memes. Such generated reasons can help content moderators better understand the severity and nature of automatically-flagged hateful memes. Similar to \cite{elsherief-2021-latent}, our work can alert the users when they intend to share a particular meme flagged as ``\textit{hateful}'' and explain the underlying reasons. This enables users to recognize the severity of the hateful meme and possibly reconsider their decision to post the meme.

\subsection{Task Definition.} 
We formulate the hateful meme explanation task as a conditional generation task dependent on meme content. Formally, given a dataset of paired hateful memes and reasons $\{x^i, r^i\}^N_{i=1}$, the goal is to learn the generation of a fluent and relevant reason conditioned on the text information $x^i_T$ and visual information $x^i_V$ extracted from the hateful meme. We can refer to the reasons as a sequence of tokens $r^i = r^i_1, \cdots, r^i_{\ell}$, where we pad the tokens to a maximal length $l$. The training objective is defined as follows:
\begin{equation}
\max_{\theta} \sum_{i=1}^{N} \sum_{j=1}^{\ell} \log{p_{\theta}(r^i_j | x^i_T, x^i_V, r^i_1, \cdots, r^i_{j-1})}
\end{equation}

where $\theta$ denotes the model’s trainable parameters.

\begin{table}[t]
\centering
\small
\begin{tabular}{c|c|c|c} 
 \hline
 & \textbf{train} & \textbf{test} & \textbf{total} \\ 
 \hline \hline
 \textbf{\#Hateful Memes} & 2,982 & 246 & 3,228 \\ 
 \hline
\end{tabular}
\caption{\textsf{HatReD} Train-Test Split}
\label{tab:dataset-distribution}
\end{table}

\subsection{Generative Models} A common model architecture used for conditional generation tasks is the encoder-decoder PLMs. Encoder-decoder PLM uses an encoder model to map the inputs to a sequence of continuous representations, which is then passed to the decoder to generate the output sequence. We train two types of PLMs in our experiments: (a) text-only PLMs that only accept text inputs; and (b) vision-language (VL) PLMs that accept text and visual inputs. 

For data pre-processing, we obtain the text information $x_T$ by tokenizing the text that overlays on the meme image. The input differences in the two types of encoder-decoder PLMs require the visual information $x_V$ to be pre-processed differently. For text-only PLMs, we extracted the the meme's image caption using ClipCap~\cite{DBLP:journals/corr/abs-2111-09734}. In addition, we applied Google Vision Web Entity Detection API and Fairface classifier~~\cite{karkkainen2019fairface} to extract the meme's entities and demographic information, respectively. Finally, we concatenate the image caption, extracted entities, and demographic information to represent the visual information. For VL PLMs, we used Detectron2 \cite{wu2019detectron2} with bottom-up attention \cite{Anderson2017up-down}\footnote{https://github.com/airsplay/py-bottom-up-attention} to extract object regions and bounding boxes from the meme's image.

\paragraph{Model Training.} We trained the PLMs on the annotated reasons  in the \textsf{HatReD} dataset. As the hateful memes in \textsf{HatReD} may contain multiple reasons, we randomly sampled one reason for each hateful meme during training. The training objective is to minimize the cross-entropy loss: 
\begin{equation}
L_{CE} = - \sum_{i=1}^{N}  \sum_{j=1}^{\ell} \log{p_{\theta}(r^i_j | x^i_T, x^i_V, r^i_1, \cdots, r^i_{j-1})}
\end{equation}

\paragraph{Model Inference.} Conditioned on the meme's text information $x_T$ and visual information $x_V$, we generate three token sequences via two following decoding strategies: (i) \textbf{greedy decoding}, which generates a sequence by greedily selecting the most probable token at each time step; and (ii) \textbf{beam search}, which generates the most likely $N$ token sequences at each time step and selecting the token sequence with the overall highest probability. We choose the token sequence with the highest overall score.

\begin{table*}[h]
\centering
\small
\begin{tabular}{ccccccccc}
\toprule
& \multicolumn{2}{c}{Models} & \multicolumn{3}{c}{N-gram matching} & \multicolumn{3}{c}{Embedding-based}  \\ 
\cmidrule(rl){2-3} \cmidrule(rl){4-6} \cmidrule(rl){7-9} 
& {Encoder} & {Decoder} & {BLEU} & {ROUGE-L} & {H. Mean} & {BERT-P} & {BERT-R} & {BERT-F} \\
\midrule
\multirow{3}{*}{Text-Only} & RoBERTa$^{base}$ & GPT2$^{base}$ & 0.068 & 0.222 & 0.104 & 0.112 & 0.327 & 0.218 \\
& RoBERTa$^{base}$ & RoBERTa$^{base}$ & 0.177 & 0.389 & 0.243 & \textbf{0.508} & 0.453 & \textbf{0.480} \\
& T5$^{large}$ & T5$^{large}$ & \textbf{0.190} & \textbf{0.392} & \textbf{0.256} & 0.485 & \textbf{0.473} & 0.479 \\ \midrule
\multirow{3}{*}{Vision-Language} & VisualBERT & GPT2$^{base}$ & 0.065 & 0.219 & 0.100 & 0.100 & 0.342 & 0.219 \\
& VisualBERT & RoBERTa$^{base}$ & 0.179 & 0.391 & 0.246 & 0.499 & 0.449 & 0.474 \\
& VL-T5 & VL-T5 & 0.180 & 0.378 & 0.244 & 0.472 & 0.409 & 0.446 \\
\bottomrule
\end{tabular}
\caption{Automatic evaluation results of the PLMs' generated reasons on \textsf{HatReD}'s test set. All metrics favor higher score and have a cap of 1. All results have a standard deviation of $\leq$ 0.03}
\label{tab:benchmark-performance}
\end{table*}

\section{Experiments}
\label{sec:experiments}

\begin{table}[t]
\centering
\small
\begin{tabular}{cccc}
\toprule
\multicolumn{2}{c}{Models} & \multicolumn{2}{c}{Metrics (Avg.)} \\
\cmidrule(rl){1-2} \cmidrule{3-4}
{Encoder} & {Decoder} & Fluency & Relevance \\
\midrule
RoBERTa$^{base}$ & GPT2$^{base}$ & 4.667 & 2.681  \\
RoBERTa$^{base}$ & RoBERTa$^{base}$ & \textbf{4.874} & 2.720  \\
T5$^{large}$ & T5$^{large}$ & 4.630 & \textbf{3.112}  \\
\midrule
VisualBERT & GPT2$^{base}$ & 4.870 & 2.283  \\
VisualBERT & RoBERTa$^{base}$ & 4.344 & 2.931  \\
VL-T5 & VL-T5 & 4.626 & 2.602  \\
\midrule
\multicolumn{2}{c}{\textit{ground truth reasons}} & \textit{4.937} & \textit{4.352}  \\
\bottomrule
\end{tabular}
\caption{Human evaluation results of the PLMs' generated reasons on \textsf{HatReD}'s test set.}
\label{tab:human-evaluation-results-hatred}
\end{table}

\subsection{Experiment Settings}
\paragraph{Baselines.} To understand the challenges of our proposed hateful meme explanation generation task, we fine-tune and evaluate encoder-decoder PLMs using the \textsf{HatReD} dataset. For text-only PLMs, we use T5 \cite{JMLR:v21:20-074}, RoBERTa \cite{liu2019roberta}, and GPT2 \cite{Radford2019LanguageMA}. As GPT2 is a decoder-only architecture, we adopt RoBERTa as its encoder. For VL PLMs, we benchmark \cite{cho2021vlt5} and VisualBERT \cite{models-li-2019-visualbert}. As VisualBERT is an encoder-only architecture, we utilized RoBERTa and GPT2 as its decoder in two different settings. The evaluation of these PLMs establishes the baselines for this new hateful meme explanation task.

\paragraph{Feature Space Alignment.} To align feature spaces between encoders and decoders with different architectures, we place the models into a sequence-to-sequence model architecture with randomly initialized cross-attention layers added to each decoder block \cite{rothe2020leveraging}. The training error, back-propagated through the cross-attention layer, fine-tunes the weights and aligns the models' feature spaces.

\paragraph{Evaluation Metrics.} We perform both automatic and human evaluations on the baselines. We adopt metrics commonly used in natural language generation tasks for automatic evaluation: (1) $N$-gram matching for word similarity; and (2) embedding-based metric for semantic similarity. For n-gram matching metrics, we compute the average BLEU~\cite{papineni-2002-bleu} and ROUGE-L~\cite{lin-2004-rouge} scores. We also compute the harmonic mean of these two metrics. We compute the precision, recall, and F1 of the BERTScore~\cite{zhang2019bertscore} for embedding-based metric. For human evaluation, we recruit human evaluators to assess the generated reasons on two aspects: \textit{fluency} and \textit{relevance}. The human evaluators are tasked to rate the generated reasons on the Likert scales described in Section~\ref{sec:data_construct}. We also mitigate positional bias by presenting the generated reasons in a scrambled order.

\begin{table}[t]
\centering
\small
\begin{tabular}{cccc}
\toprule
\multicolumn{2}{c}{Models} & \multicolumn{2}{c}{Metrics (Avg.)} \\
\cmidrule(rl){1-2} \cmidrule{3-4}
{Encoder} & {Decoder} & Fluency & Relevance \\
\midrule
T5$^{large}$ & T5$^{large}$ & \textbf{4.540} & 1.850 \\
VisualBERT & RoBERTa$^{base}$ & 3.990 & \textbf{2.040}  \\
\bottomrule
\end{tabular}
\caption{Human evaluation results of the PLMs' generated reasons on 50 hateful memes from MAMI.}
\label{tab:human-evaluation-results-mami}
\end{table}

\newcolumntype{Q}{>{\centering\arraybackslash}m{2.1cm}}
\newcolumntype{R}{>{\arraybackslash}m{6.2cm}}
\newcolumntype{S}{>{\arraybackslash}m{3.8cm}}
\newcolumntype{T}{>{\arraybackslash}m{8cm}}
\newcolumntype{U}{>{\centering\arraybackslash}m{8cm}}
\begin{table*}[ht]
  \small
  \centering
  \begin{tabular}{Q|R|S|S}
  \hline
    \multirow{2}{*}{\textbf{Hateful Meme}} & 
		\centering
    \begin{minipage}[b]{0.7\columnwidth}
		\raisebox{-.5\height}{\includegraphics[scale=0.32]{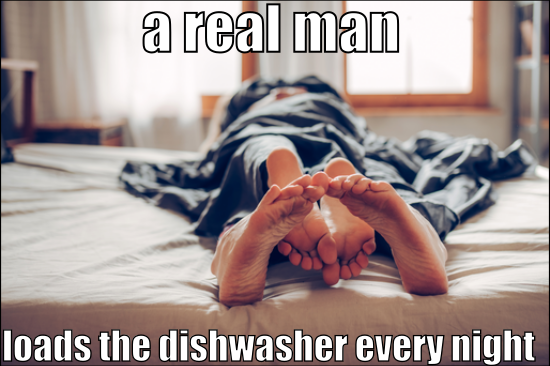}}
    \end{minipage} &
    \multicolumn{2}{c}{\centering\begin{minipage}[b]{0.7\columnwidth}
		\raisebox{-0.7\height}{\includegraphics[scale=0.22]{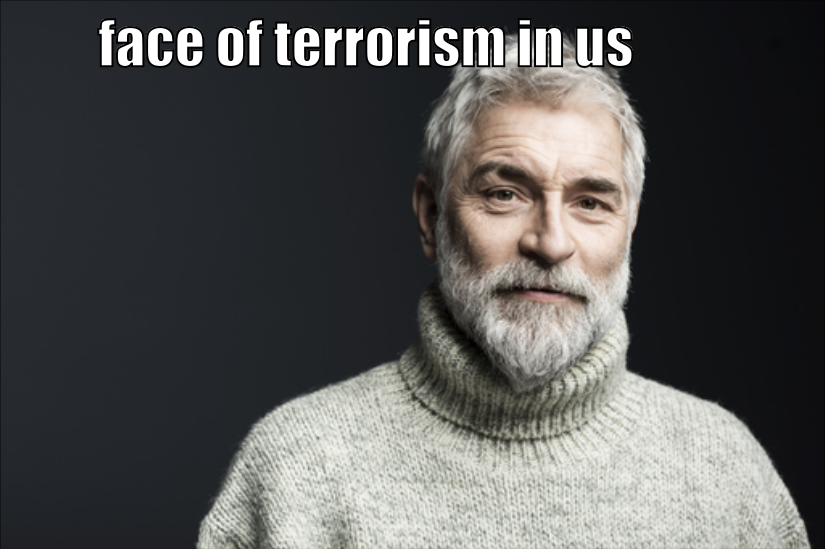}}
    \end{minipage}} \\ 
    & \centering (a) & \multicolumn{2}{U}{(b)} \\ \hline
    \textbf{Image Caption} & woman lying on a bed with her hands on her head. & portrait of a senior man. & portrait of a senior \textbf{[white]} man. \\\hline
    \textbf{T5} & dehumanizes the females as less capable humans that are only good for household chores such as \green{dishwashing} as well as \green{fulfilling the sexual needs of men} & vilifies the \textcolor{red}{muslim} by suggesting that they are terrorists & vilifies the \green{white} by suggesting that they are terrorists  \\\hline
    \textbf{VisualBERT-}\newline\textbf{RoBERTa} & dehumanizes the females as less capable humans suited for household chores like \green{dishwashing} and \green{dishwashing} & \multicolumn{2}{T}{vilifies the \textcolor{red}{immigrants} by suggesting that they are terrorists} \\\hline
    \textbf{Ground Truth} & dehumanizes the females as \green{sexual objects} as well as less capable beings only good for \green{dishwashing}.  & \multicolumn{2}{T}{ridicules the \green{whites} as terrorists by mocking the fact that the majority of shooters in the us are the whites.} \\\hline
    \end{tabular}
  \caption{Hateful memes from \textsf{HatReD} dataset with reasons generated by VisualBERT-RoBERTa and T5 models. The bracketed \textbf{[word]} in the image caption is a manual correction that explores the impact of having accurate and detailed image explanations in text-only models. The highlighted \green{green} and \red{red} words in the generated explanations outline the correct implications of hate and hallucinations (i.e. misinformation) present in the hateful memes, respectively.}
  \label{tab:qualitative-analysis-hatred}
\end{table*}

\newcolumntype{M}{>{\centering\arraybackslash}p{2.2cm}}
\newcolumntype{N}{>{\arraybackslash}m{4.6cm}}
\begin{table*}[ht]
  \small
  \begin{tabular}{M|N|N|N}
  \hline
    \multirow{2}{*}{\textbf{Hateful Meme}} & \begin{minipage}[b]{0.52\columnwidth}
		\centering
		\raisebox{-.5\height}{\includegraphics[scale=0.21]{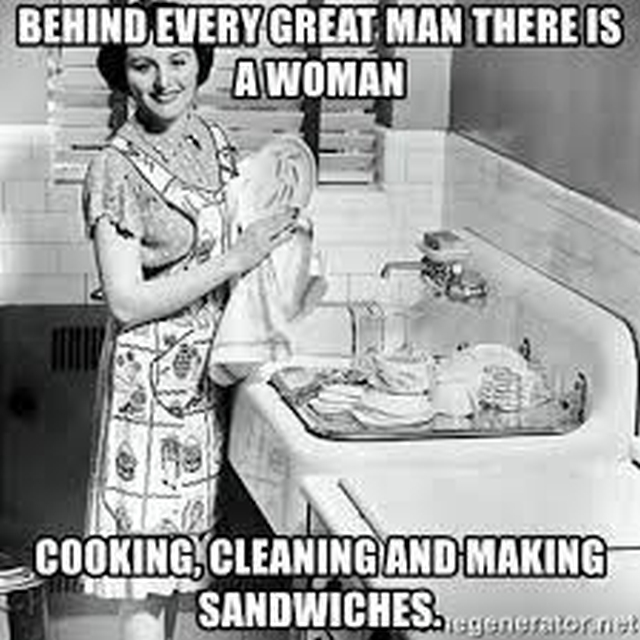}}
         \newline (a)
	\end{minipage} &
    \begin{minipage}[b]{0.52\columnwidth}
		\centering
		\raisebox{-.5\height}{\includegraphics[scale=0.27]{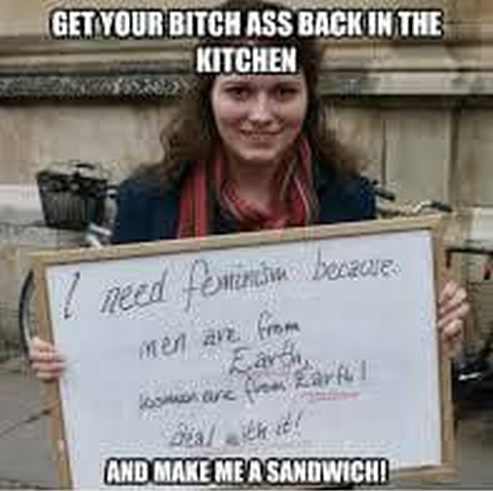}}
         \newline (b)
	\end{minipage} &
    \begin{minipage}[b]{0.52\columnwidth}
		\centering
		\raisebox{-.5\height}{\includegraphics[scale=0.21]{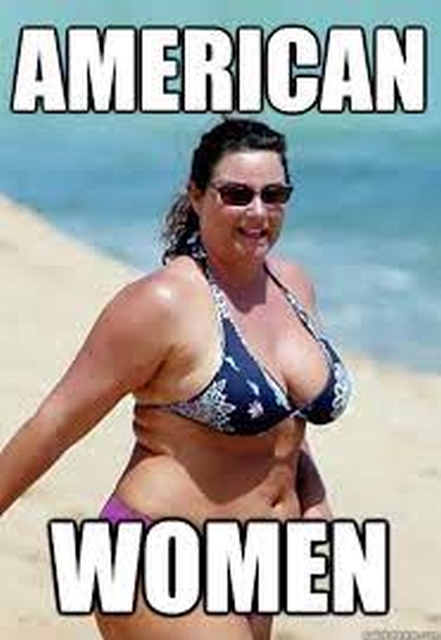}}
         \newline (c)
	\end{minipage} \\ \hline
    \textbf{VisualBERT-}\newline\textbf{RoBERTa} & dehumanizes the females as less capable humans suited for \green{household chores like cooking} & dehumanizes the females by implying that their only purpose is to \green{cook for men} & mocks the females by suggesting they are \textcolor{red}{inferior towards the white women} \\\hline
    \textbf{T5} & dehumanizes the females as less capable humans that are only good for \green{cooking, cleaning and making sandwiches} & mocks the females by implying that they are lesser people who are \green{only good for making food} & disrespects the \textcolor{red}{lgbt community} by mocking transgender women and suggesting they are \textcolor{red}{only good for sex} \\\hline
    \end{tabular}
    \caption{Hateful memes from MAMI datasets with reasons generated by VisualBERT-RoBERTa and T5 models. The highlighted \green{green} and \red{red} words in the generated explanations outline the correct implications of hate and hallucinations (i.e. misinformation) present in the hateful memes, respectively.}
  \label{tab:qualitative-analysis-mami}
\end{table*}

\subsection{Experiments on \textsf{HatReD}}
We fine-tune the baselines over ten random seeds using the \textsf{HatReD} training set and evaluate the baselines' ability to generate fluent and relevant reasons for the hateful memes in the \textsf{HatReD} test set. Table~\ref{tab:dataset-distribution} shows the distribution of the train-test split.

Table \ref{tab:benchmark-performance} shows the average automatic evaluation of the generated reasons for hateful memes in \textsf{HatReD}. We observed that T5 outperforms other PLMs across most evaluation metrics. Nevertheless, the best-performing model still performed badly with low $N$-gram matching scores (i.e., H. Mean = 0.256) and moderate BERTScore. The results suggest that the generated reasons differ substantially from the ground-truth reasons. However, the generated reasons could still convey similar meanings. Therefore, we perform human evaluations to assess the acceptability of the generated reasons.

Table \ref{tab:human-evaluation-results-hatred} shows the human evaluation results of the generated reasons and human-written reasons for hateful memes in \textsf{HatReD}. The results indicate that while most PLMs can generate fluent reasons, the generated reasons scored poorly in terms of \textit{relevance}. Specifically, T5, which has the highest average \textit{relevance} score, still had a significantly lower score compared to human-written ground-truth reasons. The superior performance of T5, a text-only model, over multimodal models can be attributed to its larger model size and the abundance of text data used for pretraining. T5 has approximately three times more parameters and is trained on 40 times more data than other models \cite{implicit-knowledge-raffel-2020-t5}. Nevertheless, the baseline PLMs' performance suggests the difficulty of the hateful meme reason generation task. We hope that the availability of \textsf{HatReD} enables researchers to design novel and better reason generation models for the task.

\begin{figure}[t]
    \centering
    \includegraphics[scale=0.37]{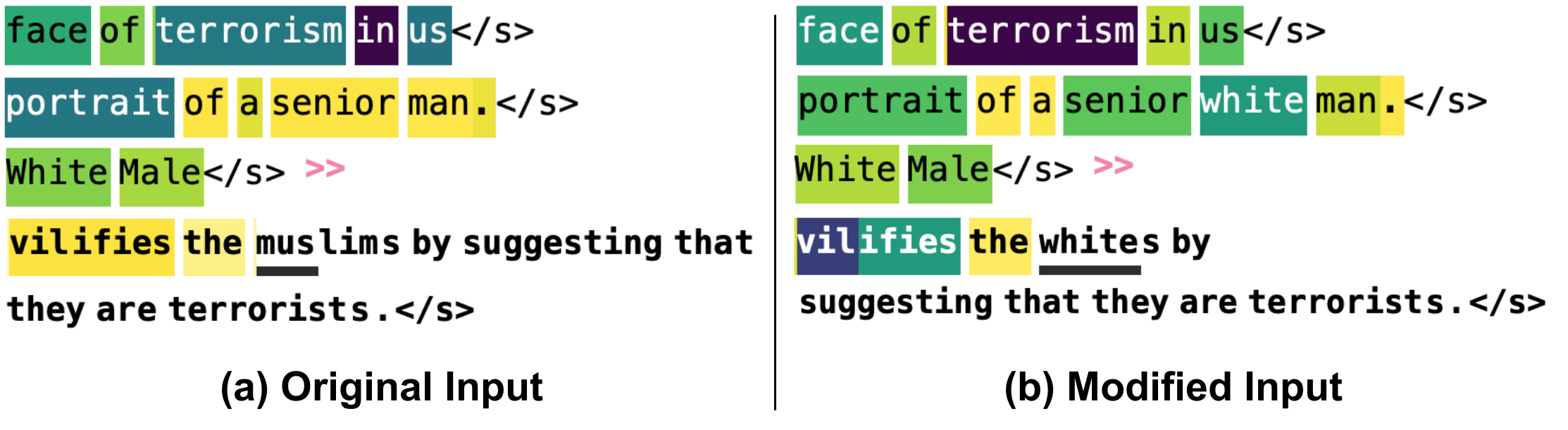}
    \caption{Input Saliency of T5 model on Meme \ref{tab:qualitative-analysis-hatred}b}
    \label{fig:input-ablation-study}
\end{figure}

\subsection{Experiment on Unseen Dataset}
To further evaluate the usefulness of \textsf{HatReD} in explaining hateful memes, we conduct a domain adaption experiment where the baselines are trained on \textsf{HatReD} and tested on an unseen dataset. Specifically, we applied two top-performance baselines, namely, VisualBERT-RoBERTa and T5, to generate the reasons for hateful memes in the Multimedia Automatic Misogyny Identification (\textbf{MAMI})\footnote{https://competitions.codalab.org/competitions/34175} datasets~\cite{fersini2022semeval}. The MAMI dataset contains hateful memes that discriminate against females. As there are no ground truth reasons for the MAMI dataset, we perform a human evaluation of the generated reasons for 50 randomly sampled MAMI hateful memes.

Table \ref{tab:human-evaluation-results-mami} shows the human evaluation results on the generated reasons for the hateful memes in MAMI. Similarly, we observe that the baselines are able to generate fluent reasons. We also noted that the generated reasons have much lower \textit{relevance} scores, which again suggests the difficulty of the hateful meme reason generation task. Nevertheless, we observe that 20\% of the hateful misogynous memes are rated highly relevant (i.e., \textit{relevancy} $\geq$ 4). The promising result demonstrates the possibility of performing domain adaption, where generative models are trained on \textsf{HatReD} to generate the reasons for hateful memes in unseen domains.


\subsection{Case Studies}

Besides performing quantitative evaluations on the hateful meme reason generation task, we also perform empirical analysis on the best performing PLM’s (i.e. T5 and VisualBERT-RoBERTa) generated reasons for hateful memes in the \textsf{HatReD} and MAMI datasets. Specifically, for this analysis, we examine generated reasons with either high (i.e., $\geq4$) or low ($<2$) relevance scores. 

Table \ref{tab:qualitative-analysis-hatred} presents examples of generated reasons for hateful memes in the \textsf{HatReD}. We found that both models generate reasons that accurately reflect the implied hate speech for the meme \ref{tab:qualitative-analysis-hatred}a. Notably, while both generated reasons reveal the intention to undermine and objectify women, only the reason generated by the text-only T5 model captures the intention to sexualize women. On the other hand, the T5 and VisualBERT-RoBERTa models generate reasons that misidentify the social target as Muslims and Immigrants for the meme \ref{tab:qualitative-analysis-hatred}b respectively. Examining the text-only T5 model that relies on extracted visual information, we observe that the image caption captures the essential visual information of a woman lying on a bed in meme \ref{tab:qualitative-analysis-hatred}a but fails to capture the crucial demographic information of the white man in meme \ref{tab:qualitative-analysis-hatred}b. The absence of demographic information in the image caption might be critical in associating the face of terrorism with white people, which led to the inaccurate identification of the social target in the generated reason for meme \ref{tab:qualitative-analysis-hatred}b. To examine this possibility, we make manual correction to the meme \ref{tab:qualitative-analysis-hatred}b’s image caption and explore the input saliency for the generated social target. We found that manually correcting the image caption helped to generate a new reason that accurately identifies the targeted social target, shown in Figure \ref{tab:qualitative-analysis-hatred}. Additionally, in Figure \ref{fig:input-ablation-study}, we explore and show the input saliency of the original and modified inputs via Integrated Gradients \cite{sundararajan2017axiomatic}. The results demonstrate that the manual correction caused the model to focus more on the words "face," "terrorism," "senior," "white," and "man," which are critical word associations required to identify the social target. These observations suggest the significance of having reliable visual information extractors to capture accurate visual information. As for the VisualBERT-RoBERTa model, the generation error is likely due to the model's inability to associate information from different modalities or understand detailed visual information such as the fact that the senior man is a white person. Nonetheless, these hallucinations demonstrate the limitation of state-of-the-art generation models and the potential for future improvement.

Table \ref{tab:qualitative-analysis-mami} showcases examples of generated reasons for hateful memes in the MAMI dataset. We observed that the generated reasons are often accurate for memes containing anti-feminism or patriarchal messages, as shown in meme \ref{tab:qualitative-analysis-mami}a and \ref{tab:qualitative-analysis-mami}b. This can be attributed to the high percentage of anti-feminism and patriarchy hateful memes in the training dataset, where approximately 26\% of the \textsf{HatReD}'s hateful memes in the female social category express anti-feminism and patriarchy messages. However, the generated reasons are found to be inaccurate when dealing with memes in new domains. For example, T5 model hallucinates that the meme \ref{tab:qualitative-analysis-mami}c implies transgender women are sexual objects, despite no indication of this in the textual or visual modalities. 

\section{Conclusion}
\label{sec:conclusion}
In this paper, we introduced \textsf{HatReD}, a new multimodal hateful memes dataset annotated with the underlying hateful contextual reasons. To the best of our knowledge, this is the first hateful meme dataset with written explanations. The availability of \textsf{HatReD} dataset opens the possibilities of training generative models to generate reasons for hateful memes, which can aid content moderators in understanding the severity and nature of flagged content. We defined a new conditional generation task to automatically explain hateful memes, and conducted extensive experiments using state-of-the-art PLMs to establish task baselines. Nevertheless, the quantitative and qualitative evaluations highlighted the difficulty of the hateful meme reason generation task. We hope that \textsf{HatReD} and our benchmark study will encourage more researchers to develop better models to generate fluent and relevant reasons for hateful memes. For future works, we aim to expand \textsf{HatReD} further to cover more domains of hateful memes. We will also explore different strategies to improve the existing reason generation model, such as using retrieval augmentation to incorporate explicit knowledge or improving the utilization of implicit knowledge in PLMs.


\section*{Ethical Statement} 
Research indicates that annotating hateful or offensive content can have negative effects. To protect our annotators, we establish three guidelines: 1) ensuring their acknowledgment of viewing potentially hateful content, 2) limiting weekly annotations and encouraging a lighter daily workload, and 3) advising them to stop if they feel overwhelmed. Finally, we regularly check in with annotators to ensure their well-being.


Another consideration is the usage of Facebook's hateful memes; users will have to agree with Facebook's usage agreement to gain access to the memes. The usage of Facebook's hateful memes in this study is in accordance with its usage agreement. Respecting Facebook's licenses on the memes, the \textsf{HatReD} dataset only contains the annotated reasons for the Facebook memes, but not the hateful memes; users will have to download the memes from the Facebook Hateful Meme challenge separately.

One of \textsf{HatReD}'s goals is to train AI systems to provide detailed warnings that explains the hateful nature of the meme content, raise users' awareness and discourages its dissemination. Nevertheless, we acknowledge the potential for malicious users to reverse-engineer and create memes that go undetected (or misunderstood) by the \textsf{HatReD}-trained AI systems. This is \textit{strongly discouraged}. In our paper, \textsf{HatReD} is utilized as training signals for post-hoc explanation generation (i.e., after the meme is flagged as hateful), not hateful meme detection. Researchers and platform providers should be cautious about including \textsf{HatReD} as training signals for hateful meme detection.

\bibliographystyle{named}
\bibliography{ijcai23}

\begin{thebibliography}{}

\bibitem[\protect\citeauthoryear{Anderson \bgroup \em et al.\egroup
  }{2018}]{Anderson2017up-down}
Peter Anderson, Xiaodong He, Chris Buehler, Damien Teney, Mark Johnson, Stephen
  Gould, and Lei Zhang.
\newblock Bottom-up and top-down attention for image captioning and visual
  question answering.
\newblock In {\em CVPR}, 2018.

\bibitem[\protect\citeauthoryear{Cao \bgroup \em et al.\egroup
  }{2022}]{cao2023prompting}
Rui Cao, Roy Ka-Wei Lee, Wen-Haw Chong, and Jing Jiang.
\newblock Prompting for multimodal hateful meme classification.
\newblock In {\em Proceedings of the 2022 Conference on Empirical Methods in
  Natural Language Processing}, pages 321--332, Abu Dhabi, United Arab
  Emirates, December 2022. Association for Computational Linguistics.

\bibitem[\protect\citeauthoryear{Cho \bgroup \em et al.\egroup
  }{2021}]{cho2021vlt5}
Jaemin Cho, Jie Lei, Hao Tan, and Mohit Bansal.
\newblock Unifying vision-and-language tasks via text generation.
\newblock In {\em ICML}, 2021.

\bibitem[\protect\citeauthoryear{ElSherief \bgroup \em et al.\egroup
  }{2021}]{elsherief-2021-latent}
Mai ElSherief, Caleb Ziems, David Muchlinski, Vaishnavi Anupindi, Jordyn
  Seybolt, Munmun De~Choudhury, and Diyi Yang.
\newblock Latent hatred: A benchmark for understanding implicit hate speech.
\newblock In {\em Proceedings of the 2021 Conference on Empirical Methods in
  Natural Language Processing}, pages 345--363, Online and Punta Cana,
  Dominican Republic, November 2021. Association for Computational Linguistics.

\bibitem[\protect\citeauthoryear{Fersini \bgroup \em et al.\egroup
  }{2022}]{fersini2022semeval}
Elisabetta Fersini, Francesca Gasparini, Giulia Rizzi, Aurora Saibene, Berta
  Chulvi, Paolo Rosso, Alyssa Lees, and Jeffrey Sorensen.
\newblock Semeval-2022 task 5: Multimedia automatic misogyny identification.
\newblock In {\em Proceedings of the 16th International Workshop on Semantic
  Evaluation (SemEval-2022). Association for Computational Linguistics}, 2022.

\bibitem[\protect\citeauthoryear{Gasparini \bgroup \em et al.\egroup
  }{2021}]{dataset-gasparini-2021-misogyny}
Francesca Gasparini, Giulia Rizzi, Aurora Saibene, and Elisabetta Fersini.
\newblock Benchmark dataset of memes with text transcriptions for automatic
  detection of multi-modal misogynistic content.
\newblock {\em arXiv preprint arXiv:2106.08409}, 2021.

\bibitem[\protect\citeauthoryear{Hee \bgroup \em et al.\egroup
  }{2022}]{2022-hee-explaining}
Ming~Shan Hee, Roy Ka-Wei Lee, and Wen-Haw Chong.
\newblock On explaining multimodal hateful meme detection models.
\newblock In {\em Proceedings of the ACM Web Conference 2022}, WWW '22, page
  3651–3655, New York, NY, USA, 2022. Association for Computing Machinery.

\bibitem[\protect\citeauthoryear{K{\"a}rkk{\"a}inen and
  Joo}{2019}]{karkkainen2019fairface}
Kimmo K{\"a}rkk{\"a}inen and Jungseock Joo.
\newblock Fairface: Face attribute dataset for balanced race, gender, and age.
\newblock {\em arXiv preprint arXiv:1908.04913}, 2019.

\bibitem[\protect\citeauthoryear{Kiela \bgroup \em et al.\egroup
  }{2020}]{kiela2020hateful}
Douwe Kiela, Hamed Firooz, Aravind Mohan, Vedanuj Goswami, Amanpreet Singh,
  Pratik Ringshia, and Davide Testuggine.
\newblock The hateful memes challenge: Detecting hate speech in multimodal
  memes.
\newblock {\em Advances in Neural Information Processing Systems},
  33:2611--2624, 2020.

\bibitem[\protect\citeauthoryear{Kiela \bgroup \em et al.\egroup
  }{2021}]{kiela21a-hatefulmemes}
Douwe Kiela, Hamed Firooz, Aravind Mohan, Vedanuj Goswami, Amanpreet Singh,
  Casey~A. Fitzpatrick, Peter Bull, Greg Lipstein, Tony Nelli, Ron Zhu, Niklas
  Muennighoff, Riza Velioglu, Jewgeni Rose, Phillip Lippe, Nithin Holla,
  Shantanu Chandra, Santhosh Rajamanickam, Georgios Antoniou, Ekaterina
  Shutova, Helen Yannakoudakis, Vlad Sandulescu, Umut Ozertem, Patrick Pantel,
  Lucia Specia, and Devi Parikh.
\newblock The hateful memes challenge: Competition report.
\newblock In Hugo~Jair Escalante and Katja Hofmann, editors, {\em Proceedings
  of the NeurIPS 2020 Competition and Demonstration Track}, volume 133 of {\em
  Proceedings of Machine Learning Research}, pages 344--360. PMLR, 06--12 Dec
  2021.

\bibitem[\protect\citeauthoryear{Lee \bgroup \em et al.\egroup
  }{2021}]{models-lee-2021-disentangling}
Roy Ka-Wei Lee, Rui Cao, Ziqing Fan, Jing Jiang, and Wen-Haw Chong.
\newblock Disentangling hate in online memes.
\newblock In {\em Proceedings of the 29th ACM International Conference on
  Multimedia}, pages 5138--5147, 2021.

\bibitem[\protect\citeauthoryear{Li \bgroup \em et al.\egroup
  }{2019}]{models-li-2019-visualbert}
Liunian~Harold Li, Mark Yatskar, Da~Yin, Cho-Jui Hsieh, and Kai-Wei Chang.
\newblock Visualbert: A simple and performant baseline for vision and language.
\newblock {\em arXiv preprint arXiv:1908.03557}, 2019.

\bibitem[\protect\citeauthoryear{Lin}{2004}]{lin-2004-rouge}
Chin-Yew Lin.
\newblock {ROUGE}: A package for automatic evaluation of summaries.
\newblock In {\em Text Summarization Branches Out}, pages 74--81, Barcelona,
  Spain, July 2004. Association for Computational Linguistics.

\bibitem[\protect\citeauthoryear{Lippe \bgroup \em et al.\egroup
  }{2020}]{lippe2020multimodal}
Phillip Lippe, Nithin Holla, Shantanu Chandra, Santhosh Rajamanickam, Georgios
  Antoniou, Ekaterina Shutova, and Helen Yannakoudakis.
\newblock A multimodal framework for the detection of hateful memes.
\newblock {\em arXiv preprint arXiv:2012.12871}, 2020.

\bibitem[\protect\citeauthoryear{Liu \bgroup \em et al.\egroup
  }{2019}]{liu2019roberta}
Yinhan Liu, Myle Ott, Naman Goyal, Jingfei Du, Mandar Joshi, Danqi Chen, Omer
  Levy, Mike Lewis, Luke Zettlemoyer, and Veselin Stoyanov.
\newblock Roberta: A robustly optimized bert pretraining approach.
\newblock {\em arXiv preprint arXiv:1907.11692}, 2019.

\bibitem[\protect\citeauthoryear{Mathias \bgroup \em et al.\egroup
  }{2021}]{dataset-mathias-2021-findings}
Lambert Mathias, Shaoliang Nie, Aida Mostafazadeh~Davani, Douwe Kiela,
  Vinodkumar Prabhakaran, Bertie Vidgen, and Zeerak Waseem.
\newblock Findings of the {WOAH} 5 shared task on fine grained hateful memes
  detection.
\newblock In {\em Proceedings of the 5th Workshop on Online Abuse and Harms
  (WOAH 2021)}, pages 201--206, Online, August 2021. Association for
  Computational Linguistics.

\bibitem[\protect\citeauthoryear{Mokady \bgroup \em et al.\egroup
  }{2021}]{DBLP:journals/corr/abs-2111-09734}
Ron Mokady, Amir Hertz, and Amit~H. Bermano.
\newblock Clipcap: {CLIP} prefix for image captioning.
\newblock {\em CoRR}, 2021.

\bibitem[\protect\citeauthoryear{Muennighoff}{2020}]{DBLP:journals/corr/abs-2012-07788}
Niklas Muennighoff.
\newblock Vilio: State-of-the-art visio-linguistic models applied to hateful
  memes.
\newblock {\em CoRR}, 2020.

\bibitem[\protect\citeauthoryear{Papineni \bgroup \em et al.\egroup
  }{2002}]{papineni-2002-bleu}
Kishore Papineni, Salim Roukos, Todd Ward, and Wei-Jing Zhu.
\newblock {B}leu: a method for automatic evaluation of machine translation.
\newblock In {\em Proceedings of the 40th Annual Meeting of the Association for
  Computational Linguistics}, pages 311--318, Philadelphia, Pennsylvania, USA,
  July 2002. Association for Computational Linguistics.

\bibitem[\protect\citeauthoryear{Pramanick \bgroup \em et al.\egroup
  }{2021a}]{dataset-pramanick-2021-harmful}
Shraman Pramanick, Dimitar Dimitrov, Rituparna Mukherjee, Shivam Sharma,
  Md.~Shad Akhtar, Preslav Nakov, and Tanmoy Chakraborty.
\newblock Detecting harmful memes and their targets.
\newblock In {\em Findings of the Association for Computational Linguistics:
  {ACL/IJCNLP}}, pages 2783--2796, 2021.

\bibitem[\protect\citeauthoryear{Pramanick \bgroup \em et al.\egroup
  }{2021b}]{DBLP:conf/emnlp/PramanickSDAN021}
Shraman Pramanick, Shivam Sharma, Dimitar Dimitrov, Md.~Shad Akhtar, Preslav
  Nakov, and Tanmoy Chakraborty.
\newblock {MOMENTA:} {A} multimodal framework for detecting harmful memes and
  their targets.
\newblock In {\em Findings of the Association for Computational Linguistics:
  {EMNLP}}, pages 4439--4455, 2021.

\bibitem[\protect\citeauthoryear{Radford \bgroup \em et al.\egroup
  }{2019}]{Radford2019LanguageMA}
Alec Radford, Jeffrey Wu, Rewon Child, David Luan, Dario Amodei, Ilya
  Sutskever, et~al.
\newblock Language models are unsupervised multitask learners.
\newblock {\em OpenAI blog}, 1(8):9, 2019.

\bibitem[\protect\citeauthoryear{Raffel \bgroup \em et al.\egroup
  }{2020}]{JMLR:v21:20-074}
Colin Raffel, Noam Shazeer, Adam Roberts, Katherine Lee, Sharan Narang, Michael
  Matena, Yanqi Zhou, Wei Li, and Peter~J. Liu.
\newblock Exploring the limits of transfer learning with a unified text-to-text
  transformer.
\newblock {\em Journal of Machine Learning Research}, 21(140):1--67, 2020.

\bibitem[\protect\citeauthoryear{Roberts \bgroup \em et al.\egroup
  }{2020}]{implicit-knowledge-raffel-2020-t5}
Adam Roberts, Colin Raffel, and Noam Shazeer.
\newblock How much knowledge can you pack into the parameters of a language
  model?
\newblock In {\em Proceedings of the 2020 Conference on Empirical Methods in
  Natural Language Processing (EMNLP)}, pages 5418--5426, Online, November
  2020. Association for Computational Linguistics.

\bibitem[\protect\citeauthoryear{Rothe \bgroup \em et al.\egroup
  }{2020}]{rothe2020leveraging}
Sascha Rothe, Shashi Narayan, and Aliaksei Severyn.
\newblock Leveraging pre-trained checkpoints for sequence generation tasks.
\newblock {\em Transactions of the Association for Computational Linguistics},
  8:264--280, 2020.

\bibitem[\protect\citeauthoryear{Sap \bgroup \em et al.\egroup
  }{2020}]{sap-2018-sbf}
Maarten Sap, Saadia Gabriel, Lianhui Qin, Dan Jurafsky, Noah~A. Smith, and
  Yejin Choi.
\newblock Social bias frames: Reasoning about social and power implications of
  language.
\newblock In {\em Proceedings of the 58th Annual Meeting of the Association for
  Computational Linguistics}, pages 5477--5490, Online, July 2020. Association
  for Computational Linguistics.

\bibitem[\protect\citeauthoryear{Sharma \bgroup \em et al.\egroup
  }{2022a}]{sharma2022disarm}
Shivam Sharma, Md~Shad Akhtar, Preslav Nakov, and Tanmoy Chakraborty.
\newblock Disarm: Detecting the victims targeted by harmful memes.
\newblock In {\em Findings of the Association for Computational Linguistics:
  NAACL 2022}, pages 1572--1588, 2022.

\bibitem[\protect\citeauthoryear{Sharma \bgroup \em et al.\egroup
  }{2022b}]{sharma2022detecting}
Shivam Sharma, Firoj Alam, Md~Akhtar, Dimitar Dimitrov, Giovanni Da~San
  Martino, Hamed Firooz, Alon Halevy, Fabrizio Silvestri, Preslav Nakov, Tanmoy
  Chakraborty, et~al.
\newblock Detecting and understanding harmful memes: A survey.
\newblock {\em arXiv preprint arXiv:2205.04274}, 2022.

\bibitem[\protect\citeauthoryear{Sharma \bgroup \em et al.\egroup
  }{2023}]{sharma2023characterizing}
Shivam Sharma, Atharva Kulkarni, Tharun Suresh, Himanshi Mathur, Preslav Nakov,
  Md~Akhtar, Tanmoy Chakraborty, et~al.
\newblock Characterizing the entities in harmful memes: Who is the hero, the
  villain, the victim?
\newblock {\em arXiv preprint arXiv:2301.11219}, 2023.

\bibitem[\protect\citeauthoryear{Sundararajan \bgroup \em et al.\egroup
  }{2017}]{sundararajan2017axiomatic}
Mukund Sundararajan, Ankur Taly, and Qiqi Yan.
\newblock Axiomatic attribution for deep networks.
\newblock In {\em International conference on machine learning}, pages
  3319--3328. PMLR, 2017.

\bibitem[\protect\citeauthoryear{Suryawanshi \bgroup \em et al.\egroup
  }{2020}]{dataset-suryawanshi-2020-multioff}
Shardul Suryawanshi, Bharathi~Raja Chakravarthi, Mihael Arcan, and Paul
  Buitelaar.
\newblock Multimodal meme dataset (multioff) for identifying offensive content
  in image and text.
\newblock In {\em Proceedings of the Second Workshop on Trolling, Aggression
  and Cyberbullying}, pages 32--41, 2020.

\bibitem[\protect\citeauthoryear{Wu \bgroup \em et al.\egroup
  }{2019}]{wu2019detectron2}
Yuxin Wu, Alexander Kirillov, Francisco Massa, Wan-Yen Lo, and Ross Girshick.
\newblock Detectron2.
\newblock \url{https://github.com/facebookresearch/detectron2}, 2019.
\newblock (accessed February 28, 2023).

\bibitem[\protect\citeauthoryear{Yang \bgroup \em et al.\egroup
  }{2022}]{yang2023crossdomain}
Chuanpeng Yang, Fuqing Zhu, Guihua Liu, Jizhong Han, and Songlin Hu.
\newblock Multimodal hate speech detection via cross-domain knowledge transfer.
\newblock In {\em Proceedings of the 30th ACM International Conference on
  Multimedia}, MM '22, page 4505–4514, New York, NY, USA, 2022. Association
  for Computing Machinery.

\bibitem[\protect\citeauthoryear{Zhang \bgroup \em et al.\egroup
  }{2019}]{zhang2019bertscore}
Tianyi Zhang, Varsha Kishore, Felix Wu, Kilian~Q Weinberger, and Yoav Artzi.
\newblock Bertscore: Evaluating text generation with bert.
\newblock {\em arXiv preprint arXiv:1904.09675}, 2019.

\end{thebibliography}

\end{document}